\documentclass[runningheads]{llncs}
\usepackage{graphicx}

\usepackage{tikz}
\usepackage{comment}
\usepackage{color}
\usepackage{multirow}
\usepackage[symbol]{footmisc}
\usepackage{graphicx}
\usepackage{amsmath}
\usepackage{amssymb}
\usepackage{booktabs}
\usepackage{float}
\usepackage{soul}
\usepackage{CJKutf8}
\usepackage{xcolor,pifont}
\usepackage{tablefootnote}
\usepackage{caption}
\usepackage{subcaption}

\usepackage[accsupp]{axessibility}  %

\usepackage[pagebackref,breaklinks,colorlinks]{hyperref}
\usepackage[width=122mm,left=12mm,paperwidth=146mm,height=193mm,top=12mm,paperheight=217mm]{geometry}

\usepackage[capitalize]{cleveref}
\crefname{section}{Sec.}{Secs.}
\Crefname{section}{Section}{Sections}
\Crefname{table}{Table}{Tables}
\crefname{table}{Tab.}{Tabs.}

\usepackage{color}
\definecolor{turquoise}{cmyk}{0.65,0,0.1,0.3}
\definecolor{purple}{rgb}{0.65,0,0.65}
\definecolor{darkgreen}{rgb}{0, 0.5, 0}
\definecolor{orange}{rgb}{0.8, 0.6, 0.2}
\definecolor{red}{rgb}{0.8, 0.2, 0.2}
\definecolor{darkred}{rgb}{0.6, 0.1, 0.05}
\definecolor{blueish}{rgb}{0.0, 0.3, .6}
\definecolor{light_gray}{rgb}{0.7, 0.7, .7}
\definecolor{pink}{rgb}{1, 0, 1}
\definecolor{greyblue}{rgb}{0.25, 0.25, 1}
\definecolor{gold}{rgb}{0.7, 0.5, 0}

\def \customparskip {0.5em}
\renewcommand{\paragraph}[1]{\vspace{\customparskip}\noindent\textbf{#1}}

\usepackage{booktabs}

\newcommand{\Figure}[1]{Figure~\ref{fig:#1}}

\newcommand{\Table}[1]{Table~\ref{tab:#1}}

\newcommand{\Eq}[1]{Eq.~\ref{eq:#1}}

\newcommand{\calF}{{\mathcal{F}}}
\newcommand{\bTheta}{{\boldsymbol{\Theta}}}
\newcommand{\bx}{{\mathbf{x}}}
\newcommand{\bc}{{\mathbf{c}}}
\newcommand{\bd}{{\mathbf{d}}}
\newcommand{\br}{{\mathbf{r}}}
\newcommand{\bC}{{\mathbf{C}}}
\newcommand{\balpha}{{\mathbf{\alpha}}}
\newcommand{\bD}{{\mathbf{D}}}
\newcommand{\Mask}{{\mathcal{M}}}
\newcommand{\loss}[1]{{\mathcal{L}_{#1}}}
\newcommand{\Transform}{{\mathcal{T}}}
\newcommand{\ErrorNet}{{\mathcal{E}}}
\newcommand{\btheta}{{\boldsymbol{\theta}}}

\newcommand{\bPhi}{{\boldsymbol{\Phi}}}

\newcommand*\colourcheck[1]{%
  \expandafter\newcommand\csname #1check\endcsname{\textcolor{#1}{\ding{52}}}%
}
\newcommand*\colourx[1]{%
  \expandafter\newcommand\csname #1x\endcsname{\textcolor{#1}{\ding{55}}}%
}
\colourcheck{darkgreen}
\colourx{red}
\newcommand{\probP}{\text{I\kern-0.15em P}}
\newcommand{\weight}{w}
\newcommand{\expo}[1]{\exp\left(#1\right)}
\newcommand{\absrp}{\sigma}
\newcommand{\deltatime}{\delta}

\begin{document}
\begin{CJK*}{UTF8}{gbsn}

\pagestyle{headings}
\mainmatter
\def\ECCVSubNumber{3737}  %

\title{NeuMan: Neural Human Radiance Field from a Single Video}

\titlerunning{NeuMan}

\author{Wei Jiang\textsuperscript{\textdagger}\inst{1,2}\and
Kwang Moo Yi\inst{1,2}\index{Yi, Kwang Moo}\and
Golnoosh Samei\inst{1}\and \\
Oncel Tuzel\inst{1}\and
Anurag Ranjan\inst{1}
}
\authorrunning{W. Jiang et al.}
\institute{Apple \and
The University of British Columbia
\\
\email{\email{\{jw221, kmyi\}@cs.ubc.ca}, \{golnoosh, otuzel, anuragr\}@apple.com}}
\maketitle

\begin{abstract}

Photorealistic rendering and reposing of humans is important for enabling augmented reality experiences.
We propose a novel framework to reconstruct the human and the scene that can be rendered with novel human poses and views from just a single in-the-wild video.
Given a video captured by a moving camera, we train two NeRF models: a human NeRF model and a scene NeRF model. To train these models, we rely on existing methods to estimate the rough geometry of the human and the scene. Those rough geometry estimates allow us to create a warping field from the observation space to the canonical pose-independent space, where we train the human model in.
Our method is able to learn subject specific details, including cloth wrinkles and accessories, from just a 10 seconds video clip, and to provide high quality renderings of the human under novel poses, from novel views, together with the background. Code will be available at \url{https://github.com/apple/ml-neuman}.

\renewcommand{\thefootnote}{\fnsymbol{footnote}}
\footnotetext[2]{Work done while interning at Apple.}

\end{abstract}

\section{Introduction}

The quality of novel view synthesis has been dramatically improved since the introduction of Neural Radiance Fields (NeRF)~\cite{mildenhall2020nerf}.
While originally proposed to reconstruct a static scene with a set of posed images, it has since been quickly extended to dynamic scenes~\cite{park2021nerfies,li2021neuralsceneflow,park2021hypernerf} and uncalibrated scenes~\cite{wang2021nerfmm,lin2021barf}.
Recent efforts also focus on animation of these radiance field models~\cite{liu2021neuralactor,peng2021neuralbody,peng2021animatable,zhang2021stnerf,Su21arxiv_A_NeRF} of human, with the aid of large controlled datasets, further extending the application domain of radiance-field-based modeling to enable augmented reality experiences.

In this work, we are interested in the scenario where only one single video is provided, and our goal is to reconstruct the human model and the static scene model, and enable novel pose rendering of the human, without any expensive multi-cameras setups or manual annotations.
However, even with the recent advancements in NeRF methods, this is far from being trivial. Existing methods~\cite{peng2021neuralbody,liu2021neuralactor} require multi-cameras setup, consistent lighting and exposure, clean backgrounds, and accurate human geometry to train the NeRF models.
As shown in Table~\ref{tab:compare}, HyperNeRF\cite{park2021hypernerf} models a dynamic scene based on a single video, but cannot be driven by human poses.
ST-NeRF~\cite{zhang2021stnerf} reconstructs each individual with a time dependent NeRF model from multiple cameras, but the editing is limited to the transformation of the bounding box.
Neural Actor~\cite{liu2021neuralactor} can generate novel poses of a human but requires multiple videos.
HumanNeRF~\cite{weng2022humannerf} builds a human model based on a single video with manually annotated masks, but doesn't show generalization to novel poses.
Vid2Actor~\cite{weng2020vid2actor} generates novel poses of a human with a model trained on a single video but cannot model the background.
We address these problems 
by introducing \emph{NeuMan}, that
reconstructs both the human and the scene with the ability to render novel human poses and novel views, from a single in-the-wild video. 
\begin{figure}[t]
    \centering
    \includegraphics[width=\textwidth]{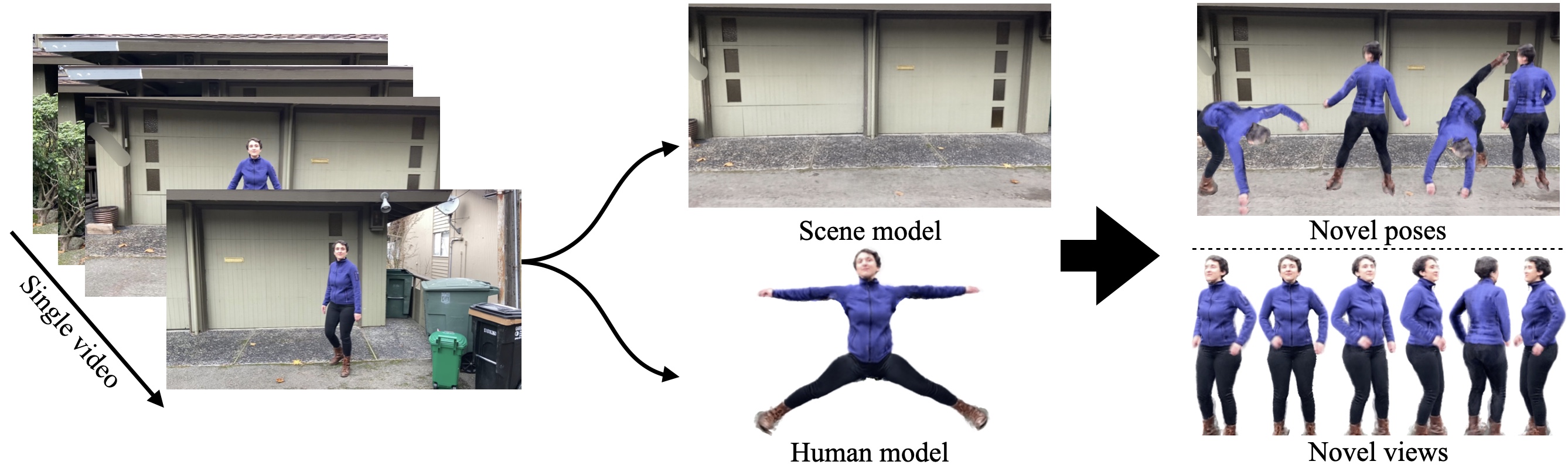}
    \caption{{\bf Teaser --}We train neural human and scene radiance fields from a single in-the-wild video to enable rendering with novel human poses and novel views.}
    \label{fig:teaser}
\end{figure}
\setlength{\tabcolsep}{8pt}

\begin{table}[t]
\begin{center}
\setlength{\tabcolsep}{3pt}
\begin{tabular}{@{}lcccc@{}}
\toprule
& \textcolor{black}{Scene Background} & \textcolor{black}{Novel Poses} & \textcolor{black}{Single Video} & \textcolor{black}{Compositionality}\\
\midrule
HyperNeRF~\cite{park2021hypernerf}    & \darkgreencheck                            & \redx                              & \darkgreencheck      & \redx                        \\
ST-NeRF~\cite{zhang2021stnerf}    & \darkgreencheck                            & \redx                              & \redx      & \darkgreencheck                        \\
Neural Actor~\cite{liu2021neuralactor} & \redx                             & \darkgreencheck                             & \redx            & \redx                    \\
HumanNeRF~\cite{weng2022humannerf}    & \redx                             & \redx                              & \darkgreencheck           & \redx                   \\
Vid2Actor~\cite{weng2020vid2actor}    & \redx                             & \darkgreencheck                             & \darkgreencheck       & \redx                       \\
\midrule
Ours         & \darkgreencheck                            & \darkgreencheck                             & \darkgreencheck           & \darkgreencheck                   \\
\bottomrule
\end{tabular}
\end{center}
\caption{
{\bf Method capacity comparison -- }
We illustrate the capabilities of existing methods. 
Compared to other methods, ours is the only one with the ability to render both the scene and the reposed human from a single video.}
\label{tab:compare}
\end{table}

\setlength{\tabcolsep}{2pt}

NeuMan is a novel framework for training NeRF models for both the human and the scene, which allows high-quality pose-driven rendering as shown in~\Figure{teaser}.
Given a video captured by a moving camera, we first estimate the human pose, human shape, human masks, as well as the camera poses, sparse scene model, and depth maps using conventional off-the-shelf methods~\cite{schoenberger2016sfm,ROMP,he2017mask,cheng2020bottom,guler2018densepose,Miangoleh2021Boosting}.

We then train two NeRF models, one for the human and one for the scene guided by the segmentation masks estimated from Mask-RCNN~\cite{he2017mask}. %
Additionally, we regularize the scene NeRF model by fusing together depth estimates from both multi-view reconstruction~\cite{schoenberger2016sfm} and monocular depth regression~\cite{Miangoleh2021Boosting}.
We train the human NeRF model in a {pose independent} canonical volume guided by a statistical human shape and pose model, SMPL~\cite{loper2015smpl} following Liu et al.~\cite{liu2021neuralactor}. We refine the SMPL estimates from ROMP~\cite{ROMP} to better serve the training. However, these refined estimates are still not perfect. Therefore, we jointly optimize the SMPL estimates together with the human NeRF model in an end-to-end fashion.
Furthermore, since our static canonical human NeRF cannot represent the dynamics that is not captured by the SMPL model, we introduce an error-correction network to counter it. The SMPL estimates and the error-correction network are jointly optimized during the training.

In summary,
\begin{itemize}
\item we propose a framework for neural rendering of a human and a scene from a single video without any extra devices or annotations;
\item we show that our method allows high quality rendering of human under novel poses, from novel views, together with the scene;
\item we introduce an end-to-end SMPL optimization and an error-correction network to enable training with erroneous estimates of the human geometry;
\item our approach allows for the composition of the human and the scene NeRF models enabling applications such as telegathering.
\end{itemize}

\section{Related Work}

As our work is mainly based on neural radiance fields, we first review works on NeRF with a focus on works that aim to control and condition the radiance fields---a necessity for rendering a human in the scene in the context of creating visual and immersive experiences~\cite{peng2021neuralbody,liu2021neuralactor}. 
We also briefly review works that aim to reanimate and perform novel view synthesis of provided scenes.

\paragraph{Neural Radiance Fields (NeRF).}
Since its first introduction~\cite{mildenhall2020nerf}, NeRF has become a popular way~\cite{dellaert2021neural} to model scenes and render them from novel views thanks to its high quality rendering.
Representing a scene as a radiance field has the advantage that \emph{by-construction} you will be able to render the scene from any supported views through volume rendering.
Efforts have been made to adapt NeRF to dynamic scenes~\cite{park2021nerfies,park2021hypernerf,li2021neuralsceneflow}, and to even edit and compose scenes with various NeRF models~\cite{yang2021objectnerf,Guo20arxiv_OSF}, widening their potential application.
While these methods have shown interesting and exciting results, they often require separate training of editable instances~\cite{Guo20arxiv_OSF} or careful curation of training data~\cite{peng2021neuralbody}.
In this work, we are interested in an \emph{in-the-wild} setup.

Particularly related to our task of interest,
various efforts have been made towards NeRF models conditioned by explicit human models, such as SMPL~\cite{loper2015smpl} or 3D skeleton~\cite{liu2021neuralactor,peng2021neuralbody,peng2021animatable,zhang2021stnerf,Su21arxiv_A_NeRF}.
Neural Body~\cite{peng2021neuralbody} associates a latent code to each SMPL vertex, and use sparse convolution to diffuse the latent code into the volume in observation space.
Neural Actor~\cite{liu2021neuralactor} learns the human in the canonical space by a volume warping based on the SMPL~\cite{loper2015smpl} mesh transformation, it also utilize a texture map to improve the final rendering quality.
Animatable NeRF~\cite{peng2021animatable} learns a blending weight field in both observation space and canonical space, and optimize for a new blending weight field for novel poses.
ST-NeRF~\cite{zhang2021stnerf} separates the human into each 3D bounding box, and learns the dynamic human within each bounding box. It doesn't require to estimate the precise human geometry, but it cannot extrapolate to unseen poses since it is dependent on time(frame).
However, all these methods require an expensive multi-cameras setup to obtain the ground truth bounding boxes, 3D poses or SMPL estimates.
In other words, they cannot be used for our purpose of reconstructing and neural rendering of human from a \emph{single} video, \emph{without} extra devices or annotations, and \emph{with} potential pose estimation errors.

HumanNeRF~\cite{weng2022humannerf}, a concurrent work, aims to create free-viewpoint rendering of human from a single video. 
While similar to our work, there are two main differences between HumanNeRF~\cite{weng2022humannerf} and ours.
First, HumanNeRF~\cite{weng2022humannerf} 
relies on manual mask annotation to separate the human from the background, while our method learns the decomposition of the human and the scene with the help of modern detectors.
Second, HumanNeRF~\cite{weng2022humannerf} 
represents motion as a combination of the skeletal and the non-rigid transformations, causing ambiguous or unknown transformations under novel poses, while ours mitigates the ambiguity by using explicit human mesh.
Another similar work is Vid2Actor~\cite{weng2020vid2actor}, which builds animatable human from a single video by learning a voxelized canonical volume and skinning weights jointly.
Although with similar goals, our method is able to reconstruct sharp human geometry with less than 40 images, comparing to thousands frames are required for Vid2Actor, our method is data-efficient.

\paragraph{Neural Rendering of Humans.}
Majority of the literature~\cite{balakrishnan2018synthesizing,ma2018disentangled,neverova2018denseposetransfer,liu2019liquid,sanyal2021learning} only consider the problem of reposing a human from a source image to a target image without changing the viewing angle which is essential for enabling new immersive experiences.
Grigorev et al~\cite{grigorev2019coordinate} tackled a similar problem to ours, namely resynthesizing a human image with a novel pose view given a single input image. They divide the problem into estimating the full texture map of the human body surface from partial texture observations and synthesizing a novel view given the estimated texture map from the first step. They employ two convolutional neural networks (CNN) for each of these steps. With the second one responsible for generating the novel pose consuming the output of the first CNN. This method does not explicitly model the source and target pose, so it is unclear how well it would perform for a novel pose.
Sarkar et al~\cite{sarkar2020neural} additionally consider re-rendering in a novel view as well as novel pose from a single input image. They use a parametric 3D human mesh to recover body pose and shape and a high dimensional UV feature map to encode appearance.

\subsection{Preliminary: Neural Radiance Fields (NeRF)}
\label{sec:preliminary}
\begin{figure}[t]
\centering
\includegraphics[width=\linewidth]{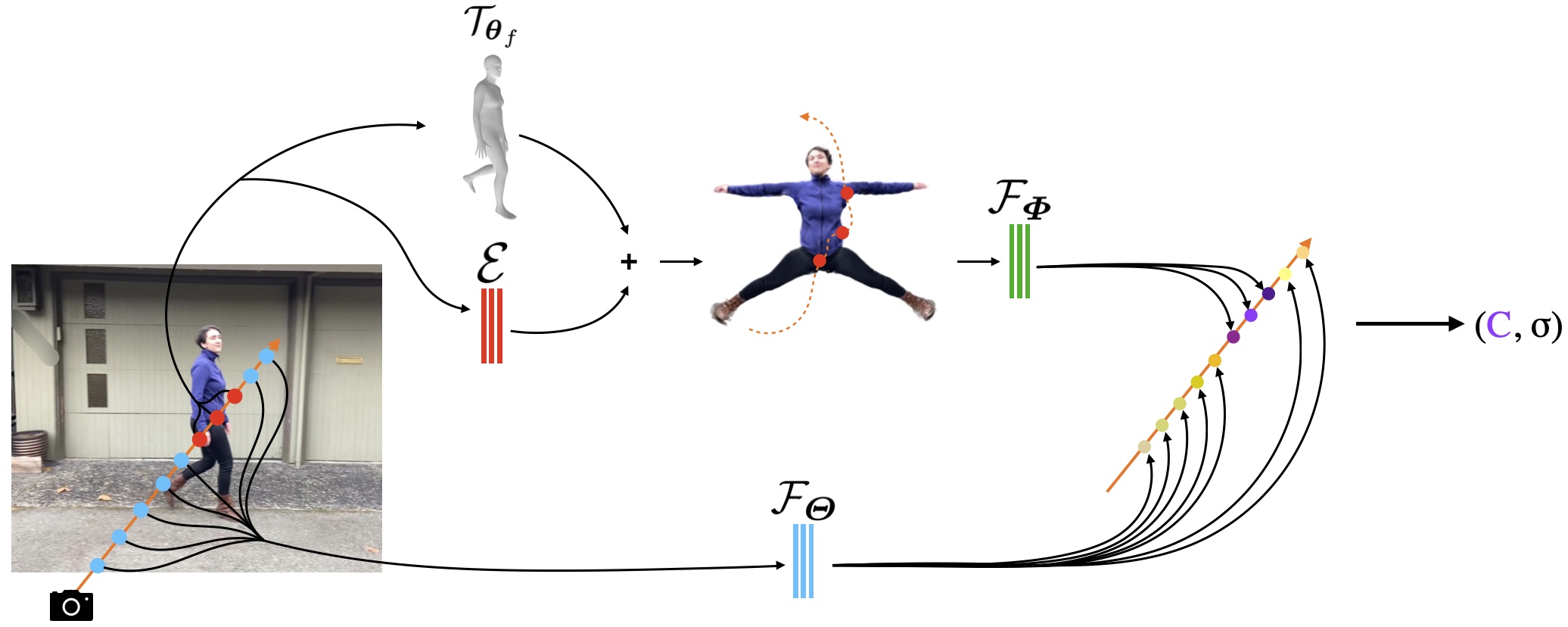}
\caption{
{\bf Overview --}
The blue samples are for the scene branch, and the red samples for the human branch. The human samples are warped to canonical space based on the estimated SMPL mesh and the error-correction network. After the RGB and opacity are evaluated, the two sets of samples are merged for the final integral to obtain the pixel color. See Section \ref{human_nerf} for more details.
}
\label{fig:overview}
\end{figure}

For completeness, we quickly review the standard NeRF model~\cite{mildenhall2020nerf}.
We denote the NeRF network as $\calF_\bTheta$ with parameters $\bTheta$, that estimates the RGB color $\bc$ and density $\sigma$ of a given a 3D location $\bx$ and a viewing direction $\bd$ as
\begin{equation}
    \bc, \sigma = \calF_\bTheta\left(\bx,\bd\right)
    \;.
\end{equation}
The radiance field function $\calF_\bTheta$ is often implemented with multi-layer perceptrons (MLPs) with positional encodings~\cite{tancik2020fourier,mildenhall2020nerf} and periodic activations~\cite{chan2021pi}.
The pixel color is then obtained 
by integrating a discretized ray $\br$ that consists of $N$ samples from the camera in the view direction $\bd$ through the volume given by
\begin{equation}
    \bC\left(\br\right) = \sum_{i=1}^{N} w_i \mathbf{c}_i
    ; \; \; \; \; \text{where, }
    w_i= \expo{- \sum_{j=1}^{i-1} \absrp_j \deltatime_j} (1-\expo{-\absrp_i \deltatime_i})
    \;, 
    \label{eq:volume_render}
\end{equation}
and $\deltatime_i$ is the distance between two adjacent samples.
Here, the accumulated alpha value of a pixel, which represents transparency, can be obtained by
$    \balpha\left(\br\right) = \sum_{i=1}^{N} w_i $.

\section{Method}

An overview of our framework is shown in \Figure{overview}. 
Our framework is composed mainly of two NeRF networks\footnote{In our work, we assume a single human being in the scene, but this can be trivially extended.}: the \textit{human} NeRF that encodes the appearance and geometry of the human in the scene, {conditioned} on the human pose; and the \textit{scene} NeRF that encodes how the background looks like.
We train the scene NeRF first, then train the human NeRF conditioned on the trained scene NeRF.

\subsection{The Scene NeRF Model}

The scene NeRF model is analogous to the background model in traditional motion detection work~\cite{sun2012layered,elgammal2000non,LIM2018256}, except it's a NeRF.
For the scene NeRF model, we construct a NeRF model and train it with only the pixels that are deemed to be from the background.

For a ray $\br$, given the human segmentation mask as $\Mask(\br)$ where $\Mask(\br)=1$ if the ray corresponds to the human and $\Mask(\br)=0$ corresponding the background, we formulate the reconstruction loss for the scene NeRF model as
\begin{equation}
    \loss{s,rgb}(\br) =  \left(1-\Mask({\bf r})\right) 
    \left\| 
        \bC_s(\br) - \hat{\bC}(\br) 
    \right\|
    \;,
    \label{eq:scenenerf}
\end{equation}
where 
$\hat{\bC}(\br)$ corresponds to the ground-truth RGB color value and $\bC_s(\br)$ corresponds to rendered color value from the scene NeRF model.

As in Video-NeRF~\cite{xian2021space}, simply minimizing \Eq{scenenerf} leads to `hazy' objects floating in the scene.
Therefore, following Video-NeRF~\cite{xian2021space},  we resolve this by adding a regularizer on the estimated density, and forcing it to be zero for space that should be empty---the space between the camera and the scene.
For each ray $\br$, we sample the terminating depth value $\hat{z}_{\br} = \bD_{fuse}(\br)$ and minimize
\begin{equation}
    \loss{s,empty}(\br) = \int_{t_n}^{\alpha\hat{z}_\br}\sigma_s\left(\br(t)\right)dt
    \;,
\end{equation}
where $\alpha=0.8$ is a slack margin to avoid strong regularization when the depth estimates are inaccurate. %
The final loss that we use to train our scene NeRF is %
\begin{equation}
    \loss{s} = \loss{s,rgb}(\br) + \lambda_{empty}\loss{s,empty}(\br)
    \;,
\end{equation}
where $\lambda_{empty}=0.1$ is a hyper parameter controlling the emptiness regularizer in all our experiments.

\paragraph{Preprocessing.} Given a video sequence, we use COLMAP~\cite{schoenberger2016mvs,schoenberger2016sfm} to obtain the camera poses,  sparse scene model, and multi-view-stereo (MVS) depth maps. 
Typically, MVS depth maps $\bD_{mvs}$ contain holes, which we fill with the help of dense monocular depth maps $\bD_{mono}$ using Miangoleh et al.~\cite{Miangoleh2021Boosting}.
We fuse $\bD_{mvs}$ and $\bD_{mono}$ together to obtain a fused depth map $\bD_{fuse}$ with consistent scale.
In more detail, 
we find a linear mapping between the two depth maps using the pixels that have both estimates.
We then transform the values of $\bD_{mono}$ with this mapping to match the depth scale in $\bD_{mvs}$ to obtain a fused depth map $\bD_{fuse}$ by filling in the holes.
For retrieving human segmentation maps we apply Mask-RCNN~\cite{he2017mask}. 
We further dilate the human masks by 4\% to ensure the human is completely masked out.
With the estimated camera poses and the background masks, we train the scene NeRF model only 
over the background.

\subsection{The Human NeRF Model} \label{human_nerf}

To build a human model that can be pose-driven, we require the model to be pose independent.
Therefore, we define a canonical space based on the 大-pose (Da-pose) SMPL~\cite{loper2015smpl} mesh, similar to~\cite{peng2021animatable,liu2021neuralactor}.
In comparison to the traditional T-pose, Da-pose avoids volume collision when warping from observation space to canonical space for the legs.

To render a pixel of a human in the observation space with this model, we transform the points along that ray into the canonical space. The difficulty in doing so is how one expands the transformation of SMPL meshes into the entire observation space to allow this ray tracing in canonical space.
Similar to Liu et al.~\cite{liu2021neuralactor}, we use a simple strategy to extend the mesh skinning into a volume warping field.

\begin{figure}
\centering
\includegraphics[width=\linewidth]{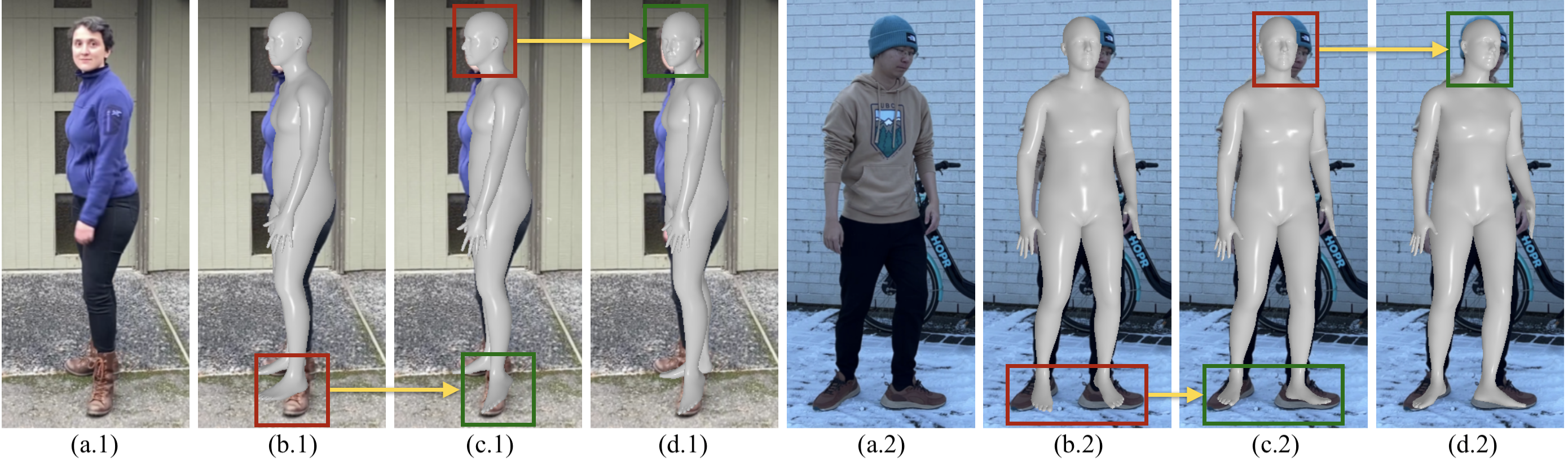}
\caption{
\textbf{Pose optimization examples--}
Our proposed preprosceeing and end-to-end optimization over SMPL parameters effectively produce better fits to the human, see the bounding boxes.
(a) Original image.
(b) ROMP estimates.
(c) Preprocessed SMPL.
(d) End-to-end optimized SMPL.
} %
\label{fig:pose_optim}
\end{figure}

In each frame $f$, given a 3D point $\bx_f=\br_f(t)$ in observation space and the corresponding estimated SMPL mesh $\btheta_f$ obtained from preprocessing~\ref{smpl_preprocess}, we transform it into the canonical space by following the rigid transformation of its closest point on the mesh; see~\Figure{overview}. 
We denote this mesh-based transform as $\Transform$ such that
$    \bx_f' = \Transform_{\btheta_f}\left(\bx_f\right)$.
This transformation, however, relies completely on the accuracy of $\btheta_f$, which is not reliable even with the recent state of the art. 
To mitigate the misalignment between the SMPL estimates and the underlying human, we propose to jointly optimize $\btheta_f$ together with the neural radiance field while training.
In Figure~\ref{fig:pose_optim}, we show the effect of online optimization on correcting the estimates. %
Furthermore, to account for the details that can not be expressed by the SMPL model, we introduce the error-correction network $\ErrorNet$, an MLP that corrects for the errors in the warping field.
Finally, the mapping between the points in the observation space to the corrected points in the canonical space $ \bx_f \rightarrow \tilde{\bx}_f'$ is obtained as
\begin{equation}
    \tilde{\bx}_f' = \Transform_{\btheta_f}\left(\bx_f\right) + \ErrorNet\left(\bx_f, f\right)
    \;.
\end{equation}
The error-correction net is only used during training, and is discarded for rendering with validation and novel poses.
Since a single canonical space is used to explain all poses, the error-correction network naturally overfits to each frame and makes the canonical volume more generalized.

Due to the nature of the warping field, a straight line in the observation space is curved in the canonical space after warping. 
Therefore, we recompute the viewing angles by taking into account how the light rays \textit{actually} travel in the canonical space by looking at where the previous sample is,
\begin{equation}
    \bd(t_i)_f' = \frac{\tilde{\bx}_f'(t_i) - \tilde{\bx}_f'(t_{i-1})}{\left\| \tilde{\bx}_f'(t_i) - \tilde{\bx}_f'(t_{i-1}) \right\|}
    \;,
\end{equation}
where $\tilde{\bx}_f'(t_i)$ and $\bd(t_i)_f'$ are the coordinate and viewing angel of the $i$-th sample along the curved ray in canonical space.
Finally, with the canonical space coordinates $\tilde{\bx}_f'$ and the corrected viewing angles $\bd_f'$ the radiance field values for the human model is obtained by 
\begin{equation}
    \bc_h, \sigma_h = \calF_\bPhi\left(\tilde{\bx}_f', \bd_f'\right)
    \;,
    \label{eq:humannerf_cs}
\end{equation}
where $\bPhi$ are the parameters of the human NeRF $\calF_\bPhi$.

To render a pixel, we shoot two rays, one for the human NeRF, and the other for the scene NeRF.
We evaluate the colors and densities for the two sets of samples along the rays. We then sort the colors and the densities in the ascending order based on their depth values, similar to ST-NeRF~\cite{zhang2021stnerf}. Finally, we integrate over these values to obtain the pixel using Eq.~\eqref{eq:volume_render}.

\paragraph{Training.} \label{training}
To train the human radiance field, we sample rays on the regions covered by the human mask
and minimize
\begin{equation}
    \loss{h,rgb}(\br) =  \Mask({\bf r})
    \left\| 
        \bC_h(\br) - \hat{\bC}(\br) 
    \right\|
    \;,
    \label{eq:human_rgb}
\end{equation}
where $\bC_h(\br)$ is the rendered color from the human NeRF model.
Similar to HumanNeRF~\cite{weng2022humannerf}, we also use LPIPS~\cite{zhang2018perceptual} as an additional loss term $\loss{lpips}$ by sampling a $32 \times 32$ patch. %
We use $\loss{mask}$ to enforce the accumulated alpha map from the human NeRF to be similar to the detected human mask.
\begin{equation}
    \loss{mask}(\br) =  \Mask({\bf r})
    \left\| 
        1 - \balpha_h(\br)
    \right\|
    \;,
    \label{eq:human_mask}
\end{equation}
where $\balpha_h$ corresponds to accumulated density over the ray as defined in Sec. \ref{sec:preliminary}.

To avoid blobs in the canonical space and semi-transparent canonical human, we enforce the volume inside the canonical SMPL mesh to be solid, while enforcing the volume outside the canonical SMPL mesh to be empty, given by
\begin{equation}
    \loss{smpl}(\hat{\bx}_f', \sigma_h) =  
    \begin{cases}
        \left\| 1 - \sigma_h \right\|,               & \text{if } \hat{\bx}_f' \text{ inside SMPL mesh}\\
        \lvert \sigma_h \rvert,              & \text{otherwise}
    \end{cases}
    \;,
    \label{eq:human_smp}
\end{equation}

Moreover, we utilize hard surface loss $\loss{hard}$~\cite{rebain2022lolnerf} to mitigate the halo around the canonical human. To be specific, we encourage the weight of each sample to be either 1 or 0 given by,
\begin{equation}
    \loss{hard} = -\log(e^{-|\weight|} + e^{-|1-\weight|})
    \label{eqn:huamn_hard}
\end{equation}
where $w$ refers to the transparency where the ray terminates as defined in Sec.~\ref{sec:preliminary}.
However, this penalty alone is not enough to obtain a sharp canonical shape, we also add a canonical edge loss, $\loss{edge}$. By rendering a random straight ray in the canonical volume, we encourage the accumulated alpha values to be either 1 or 0. This is given by,
\begin{equation}
    \loss{edge} = -\log(e^{-|\balpha_c|} + e^{-|1-\balpha_c|})
    \label{eqn:huamn_edge}
\end{equation}
where $\balpha_c$ is the accumulated alpha value obtained from a random straight ray in canonical space.
Thus, the final loss is given by,
\begin{equation}
\begin{split}
    \loss{} = \loss{h,rgb} + \lambda_{lpips}\loss{lpips} + \lambda_{mask}\loss{mask} \\
    \; \; + \lambda_{smpl}\loss{smpl} + \lambda_{hard}\loss{hard} + \lambda_{edge}\loss{edge}.
\end{split}
\end{equation}

To train, we jointly optimize $\btheta_f$, $\ErrorNet$, and $\calF_\bPhi$ by minimizing this loss. We set $\lambda_{lpips}=0.01$, $\lambda_{mask}=0.01$, $\lambda_{smpl}=1.0$, $\lambda_{hard}=0.1$, and $\lambda_{edge}=0.1$. Since the detected masks are inaccurate, we linearly decay $\lambda_{mask}$ to 0 through the training.

\paragraph{Preprocessing.}  \label{smpl_preprocess}
We utilize ROMP~\cite{ROMP} to estimate the SMPL~\cite{loper2015smpl} parameters of the human in the videos. However, the estimated SMPL parameters are not accurate. Therefore, we refine the SMPL estimates by optimizing the SMPL parameters using silhouette estimated from~\cite{guler2018densepose,wu2019detectron2}, and 2D joints estimated from~\cite{cheng2020bottom,mmpose2020} as detailed in the supplementary material. We then align the SMPL estimates in the scene coordinates. 

\begin{figure}[H]
\centering
\includegraphics[width=\linewidth]{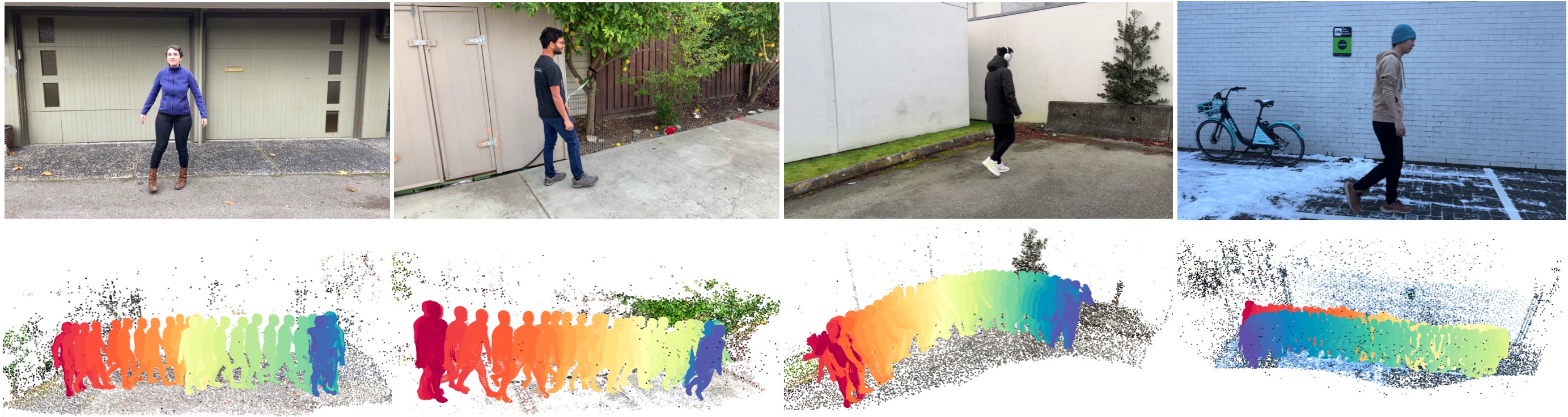}
\caption{
\textbf{Visualization of SMPL and scene alignment--}
We show sampled video frames (first row), and the estimated SMPL meshes overlaying on top of the scene point cloud (second row). The human is in the scene with a correct scale as the foot is touching the ground plane. SMPL meshes are colored based on time.
} %
\label{fig:alignment}
\end{figure}
 
\paragraph{Scene-SMPL Alignment.}
To compose a scene with a human in novel view and pose, and to train the two NeRF models, we align the coordinate systems in which the two NeRF models lie.
This is, in fact, a non-trivial problem, as human body pose estimators~\cite{kocabas2020vibe,ROMP,lin2021metro} operate in their own camera systems with often near-orthographic camera models.
To deal with this issue, we first solve the Perspective-n-Point (PnP) problem~\cite{Lepetit:160138} between the estimated 3D joints and the projected 2D joints with the camera intrinsics from COLMAP.
This solves the alignment up to an arbitrary scale.
We then assume that the human is standing on a ground at least in one frame, and solve for the scale ambiguity by finding the scale that allows the feet meshes of the SMPL model to touch the ground plan. We obtain the ground plane by applying RANSAC. We show the results of the aligned SMPL estimates in the scene in Figure~\ref{fig:alignment}.

Once the two NeRF models are properly aligned we can render the pixel by shooting two rays, one for the human NeRF model, and the other for the scene NeRF model, as describe above.
For the near and far planes to generate samples in \Eq{volume_render}, we use the estimated scene point cloud to determine them for scene NeRF, and use the estimated SMPL mesh to determine them for the human NeRF, following the strategy in Liu et al.~\cite{liu2021neuralactor}.

\begin{figure}[!b]
\centering
\includegraphics[width=\linewidth]{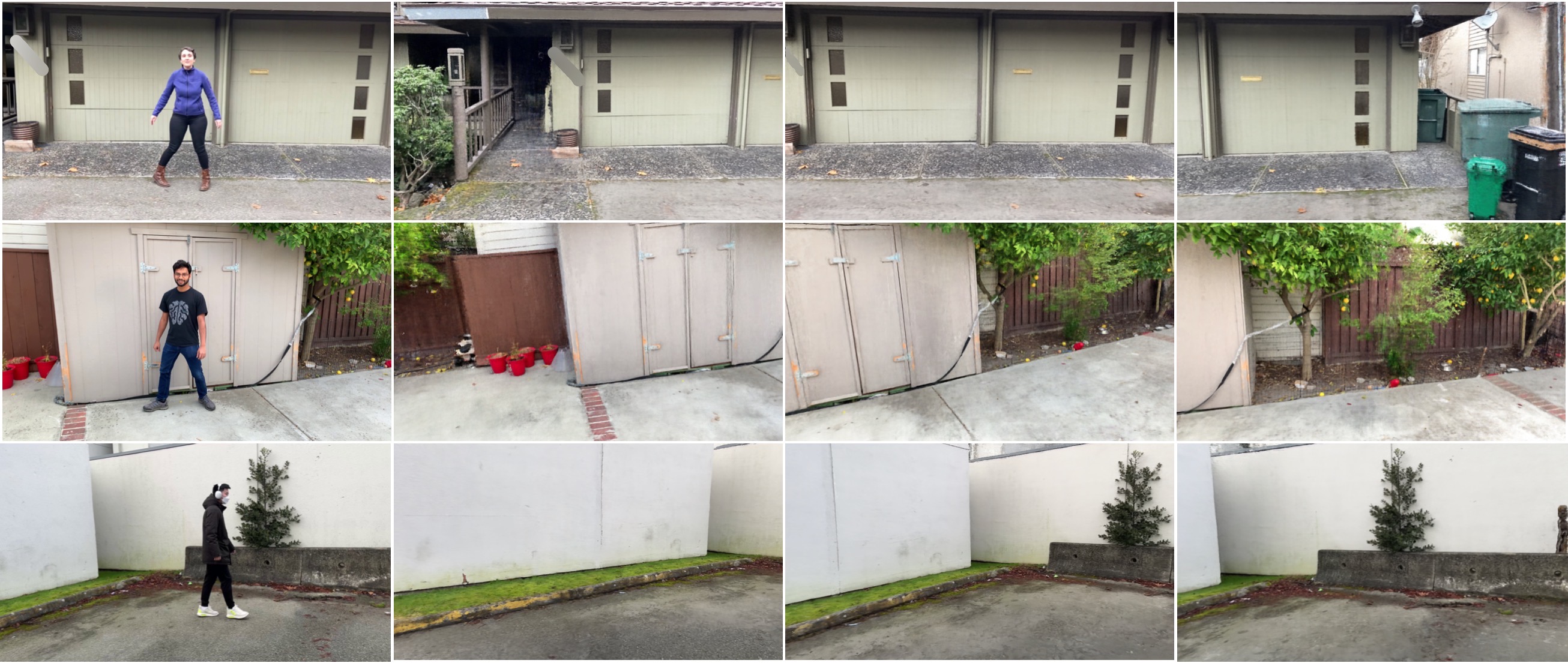}
\caption{
{\bf Scene NeRF examples -- }
We show the training samples (left), together with the renderings from validation views. Our scene models are able to remove the human in the scene effectively, and to produce high quality novel view renderings of the background even with limited coverage of the scene.
} %
\label{fig:background}
\end{figure}
 
\section{Experiments}
We introduce our dataset, show qualitative and quantitative results of our method and provide ablation studies. Our method is the first method that can render a human together with a scene with novel human poses and novel views from a single video. We show the importance of our geometry correction and novel loss terms in order to obtain a realistic and sharp human NeRF model.

\subsection{Dataset}
Existing human motion datasets are not suitable for our experiments. Generally, motion capture data~\cite{peng2021neuralbody,h36m_pami} is captured with a static multi-cameras system in a controlled environment which defeats the purpose of reconstructing from a single video. Other video datasets~\cite{milan2016mot16,ranjan2020learning} have multiple humans and crowded scenes that are not suitable for our case. Therefore, we introduce NeuMan dataset, a collection of 6 videos about 10 to 20 seconds long each, where a single person performs a walking sequence captured using a mobile phone. Moreover, the camera reasonably pans through the scene to enable multi-view reconstruction. The sequences are named -- Seattle, Citron, Parking, Bike, Jogging and Lab.
For each video sequence, we first subsample the frames from the video and use them to train our NeRF models. We split frames into 80\% training frames, 10\% validation frames, and 10\% test frames. We provide the details in supplementary material.

\begin{figure}[!b]
\centering
\includegraphics[width=0.9\linewidth]{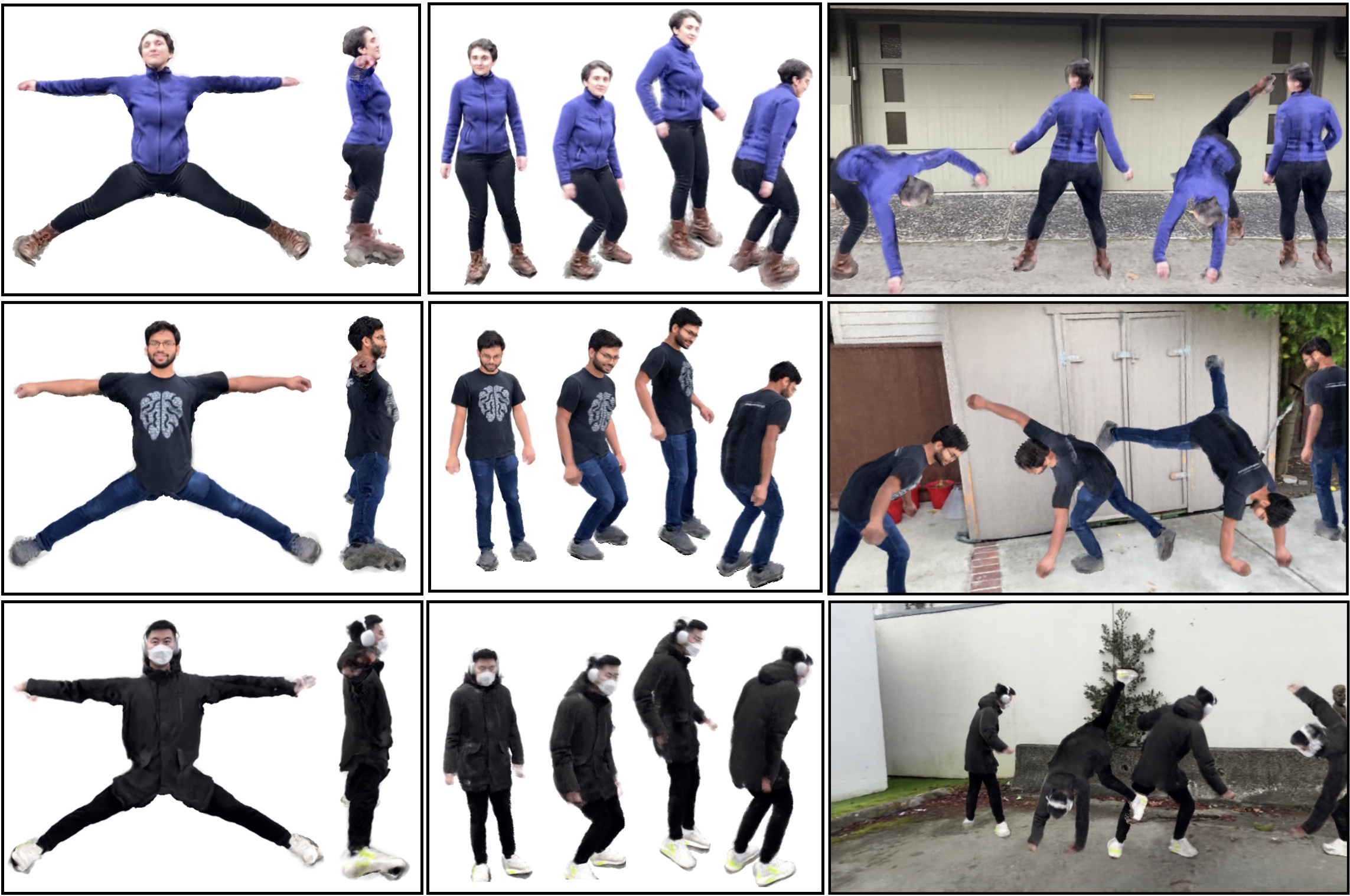}
\caption{
{\bf Human NeRF model examples -- }
We show front and side view of the reconstructed canonical human (left), novel human poses and views renderings with the reconstructed human model (middle), and composition of both the human and scene model with novel human poses (right).
} %
\label{fig:human}
\end{figure}

\subsection{Qualitative Results}
\paragraph{Scene NeRF Reconstructions.} 
Figure~\ref{fig:background} shows the novel view renderings of our scene NeRF models. By reconstructing the background pixels only, our model learns the consistent geometry of the scene, and effectively removes the dynamic human.

\paragraph{Human NeRF Reconstructions.}
Our framework learns an animatable human model with realistic details.
It captures not only the texture details such as the pattern on the cloth, but also the subject specific geometric details, such as the sleeves, collar, even zipper. Notice that these geometric details are beyond the expressiveness of SMPL model.
The learned human models can be reposed to novel driving motions, and produce high quality rendering of the human under novel poses from novel views. Although, the training sequence is as simple as walking, the model can perform stunning cartwheeling motion, which shows the ability to extrapolate to unseen poses.
By simple composition, we can render both the human and the scene realistically; see Figure~\ref{fig:human}.

\begin{figure}
\centering
\includegraphics[width=0.9\linewidth]{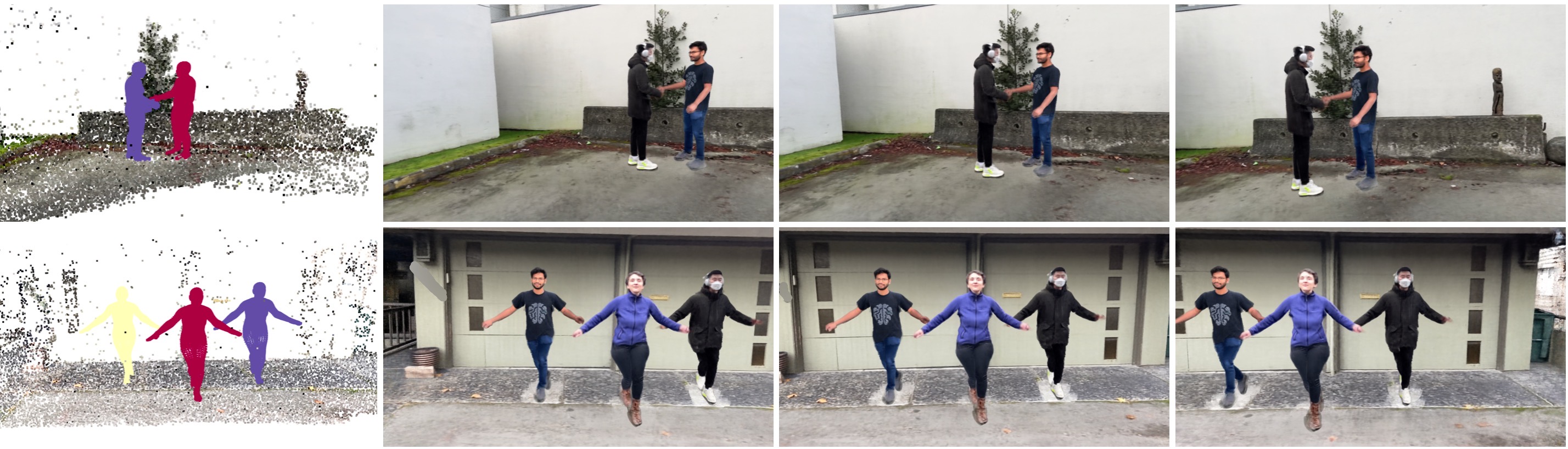}
\caption{
\textbf{Telegathering--}
Our method is able to provide telegathering for multiple individuals: (left) is the SMPL meshes and scene point cloud overlay, others are the rendering results from novel views. 
} %
\label{fig:gathering}
\end{figure}

\paragraph{Telegathering.}
The ability to render reposed human together with the scene further allows us to, for example, create telegathering of multiple individuals as in Figure \ref{fig:gathering}. The results show that our framework can facilitate combining human NeRF models in the same scene without any additional training.

\begin{figure}[H]
\centering
\includegraphics[width=0.9\linewidth]{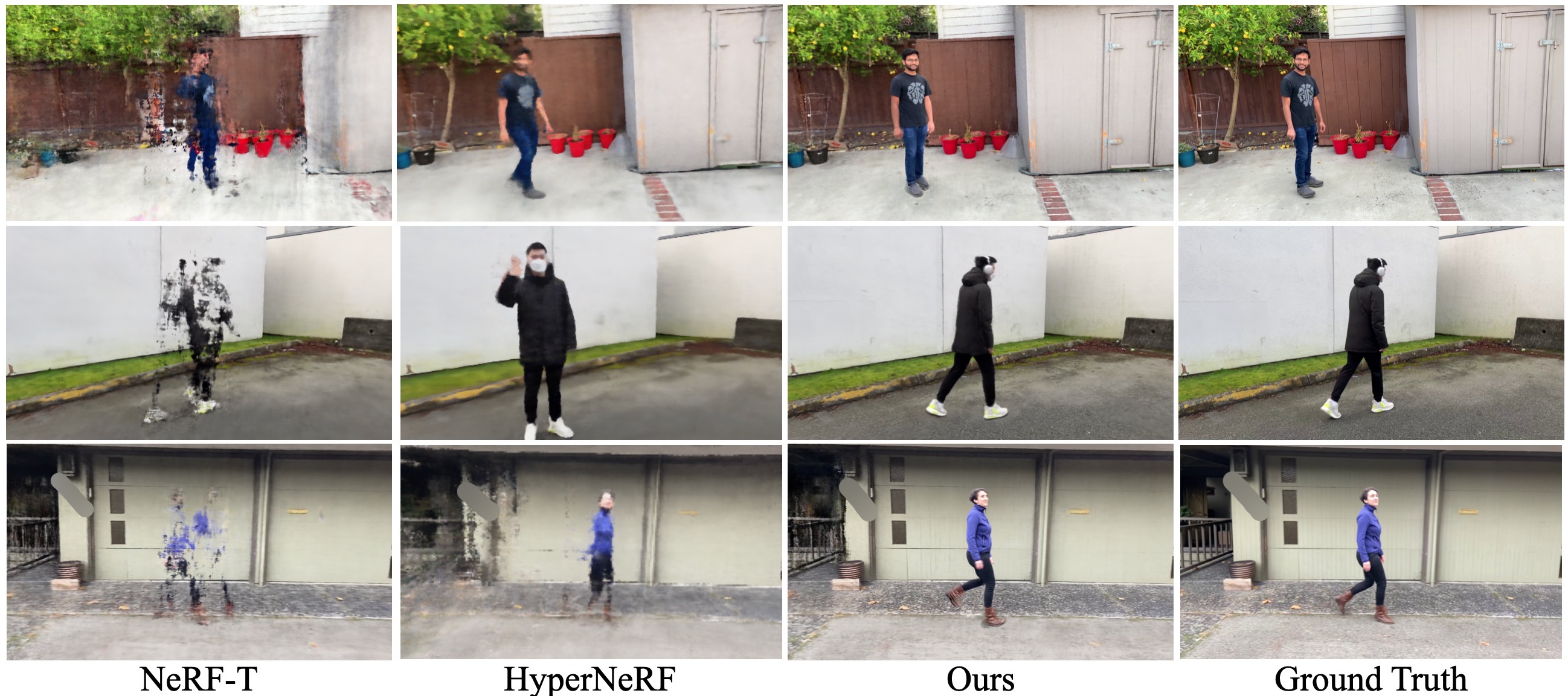}
\caption{
\textbf{Qualitative comparisons--}
We show the qualitative comparisons among NeRF-T, HyperNeRF and ours. NeRF-T and HyperNeRF failed to reconstruct the drastically dynamic scenes and to interpolate across space and time, while our method faithfully reconstructed both the human and the scene with high rendering quality.
} %
\label{fig:qualitative}
\end{figure}

\subsection{Novel view synthesis}
To compare our method to existing solutions that can do similar tasks, we train NeRF with time(NeRF-T)~\cite{li2021neuralsceneflow} using the same training data as ours but without the empty space penalty. As HyperNeRF~\cite{park2021hypernerf} requires a smoothly changing video, we provide it with densely sampled frames from the videos.
NeRF-T and HyperNeRF~\cite{park2021hypernerf} do not perform well at reconstruction of drastically dynamic scenes with humans, while our method is able to do so even with less than 40 training images.
We evaluate methods on the test views, and measure PSNR, SSIM~\cite{ssim}, and LPIPS~\cite{zhang2018perceptual}. The results are shown in~\Table{novel_view} and ~\Figure{qualitative}. Our method outperforms across all scenes and metrics.
Notice that, unlike HyperNeRF, our method depends on the estimated human pose, and we discard the offset network when rendering validation or test views. Therefore, our reported numbers also reflect the errors in the estimated human pose.

\begin{table}[t]
\begin{center}
\begin{tabular}{lccccccccc}
\toprule
          & \multicolumn{3}{c}{\begin{tabular}[c]{@{}c@{}}Seattle\end{tabular}}  & \multicolumn{3}{c}{\begin{tabular}[c]{@{}c@{}}Citron\end{tabular}}   & \multicolumn{3}{c}{\begin{tabular}[c]{@{}c@{}}Parking\end{tabular}} \\
          & \multicolumn{1}{c}{PSNR$\uparrow$} & \multicolumn{1}{c}{SSIM$\uparrow$} & \multicolumn{1}{c}{LPIPS$\downarrow$} & \multicolumn{1}{c}{PSNR$\uparrow$} & \multicolumn{1}{c}{SSIM$\uparrow$} & \multicolumn{1}{c}{LPIPS$\downarrow$} & \multicolumn{1}{c}{PSNR$\uparrow$} & \multicolumn{1}{c}{SSIM$\uparrow$} & \multicolumn{1}{c}{LPIPS$\downarrow$} \\
\midrule
\small
NeRF-T &        21.84       &      0.69              &       0.37           &               12.33  &          0.49            &        0.65             &       21.98            &             0.69        &        0.46            \\
\small
HyperNeRF &    16.43            &     0.43         &    0.40            &     16.81            &    0.41        &          0.56        &      16.04           &           0.38      &         0.62           \\
\midrule
\small
Ours      &        \bf{23.98}                  &       \bf{0.77}              &                \bf{0.26}      &   \bf{24.71}            &           \bf{0.80}               &       \bf{0.26}                    &      \bf{25.43}          &            \bf{0.79}              &       \bf{0.31}              \\
\bottomrule
\\ 
\end{tabular}

\begin{tabular}{lccccccccc}
\toprule
          & \multicolumn{3}{c}{\begin{tabular}[c]{@{}c@{}}Bike\end{tabular}}  & \multicolumn{3}{c}{\begin{tabular}[c]{@{}c@{}}Jogging\end{tabular}}   & \multicolumn{3}{c}{\begin{tabular}[c]{@{}c@{}}Lab\end{tabular}} \\
          & \multicolumn{1}{c}{PSNR$\uparrow$} & \multicolumn{1}{c}{SSIM$\uparrow$} & \multicolumn{1}{c}{LPIPS$\downarrow$} & \multicolumn{1}{c}{PSNR$\uparrow$} & \multicolumn{1}{c}{SSIM$\uparrow$} & \multicolumn{1}{c}{LPIPS$\downarrow$} & \multicolumn{1}{c}{PSNR$\uparrow$} & \multicolumn{1}{c}{SSIM$\uparrow$} & \multicolumn{1}{c}{LPIPS$\downarrow$} \\
\midrule
\small
NeRF-T &   21.16     &      0.71         &     0.36         &     20.63        &       0.53         &         0.49        &     20.52         &    0.75      &       0.39        \\
\small
HyperNeRF &     17.64     &   0.42   &      0.43  &       18.52          &  0.39    & 0.52               &      16.75      &    0.51      &     0.23          \\
\midrule
\small
Ours      &     \bf{25.52}          &     \bf{0.82}        &      \bf{0.23}         &     \bf{22.68}    &          \bf{0.67}      &         \bf{0.32}          &   \bf{24.93}      &        \bf{0.85}          &       \bf{0.21}        \\
\bottomrule
\end{tabular}

\end{center}
\caption{
{\bf Novel view synthesis comparison --} We report the quantitative results on test views.
Our method achieves the best rendering quality across all scenes and all metrics by a large margin.
Notice that the numbers for ours method depends on the estimated human pose.
}
\label{tab:novel_view}
\end{table}

\begin{figure}
\centering
\includegraphics[width=\linewidth]{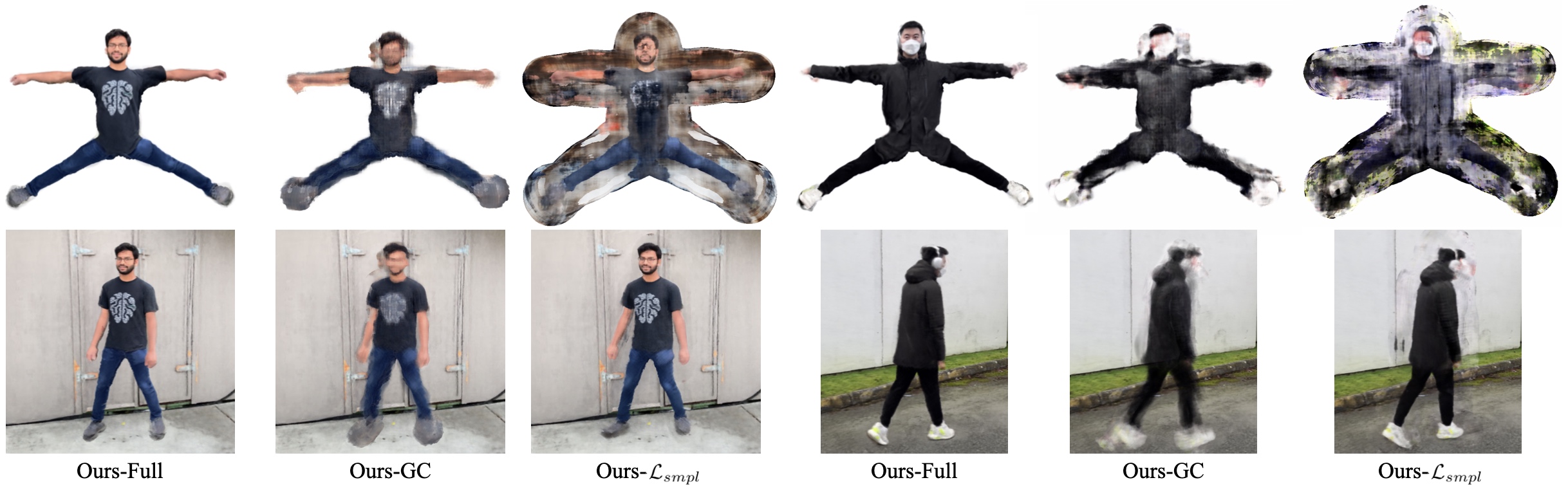}
\caption{
{\bf Ablation studies --}
We show the canonical (first row) and test view (second row) renderings of our full model, ours without any geometry corrections, and ours without $\mathcal{L}_{smpl}$. Geometry correction allows us to reconstruct the human with details, and $\mathcal{L}_{smpl}$ encourages a clean canonical volume and suppresses halo in the final renderings.
} %
\label{fig:ablation_fig}
\end{figure}
 
\subsection{Ablation Studies}
\paragraph{Effect of Geometry Correction}
Our method has 3 components to correct the estimated geometry of the human: the offline SMPL optimization in the preprocessing, the online end-to-end SMPL optimization during the training, and the error-correction network which accounts for the warping errors.
Without any geometry corrections components(Ours-GC), the canonical volume overfits to each observations causing averaged color and shape instead of sharp details of the human.
However, with geometry correction enabled, the human NeRF model can learn both the textural and geometric details of the human.
We show the comparison between Ours-Full and Ours-GC in~\Figure{ablation_fig}.

\paragraph{Effect of $\loss{smpl}$}
When $\loss{smpl}$ is disabled, fog appears in the canonical volume around the human, and causes a halo in the final renderings, see Ours-$\loss{smpl}$ in~\Figure{ablation_fig}. By encouraging the volume outside the canonical SMPL mesh to be empty, $\loss{smpl}$ effectively removes the fog and suppresses the halos.
\section{Conclusions}

We have proposed a novel framework to reconstruct the human and the scene NeRF models that can be rendered with novel human poses and views from a single in-the-wild video.
To do so, we use off-the-shelf methods to estimate either 2D or 3D geometry of the scene and the human to provide initialization.
Our human NeRF model is able to learn texture details such as patterns on cloth, and geometric details such as sleeves, collar even zipper from less than 40 images.

\paragraph{Limitations and future work.}
One major limitation is that the dynamics beyond SMPL cannot be modeled with our static NeRF, those dynamics will degenerate to average shape or color.
This is most evident in the hands when the gestures are changing over frames, see Figure~\ref{fig:human}.
We hope to apply more expressive body models to model hand gestures~\cite{smplx}, even garment dynamics~\cite{ma2020cape}, to mitigate this problem.

Additionally, our warping function is a simple extension of the SMPL mesh skinning, it could cause volume collision in some extreme cases.
A collision-aware volume warping method or a learned one is required to improve the generalization under extreme poses.

Finally, we assume that the human always has at least one contact point with the ground to estimate the scale relative to the scene.
To apply our method to videos with jumping or uneven ground, we need smarter geometric reasoning.

\paragraph{Acknowledgement}
We thank Ashish Shrivastava, Russ Webb and Miguel Angel Bautista Martin for providing insightful review feedback.
\bibliographystyle{splncs04}
\bibliography{egbib}

\clearpage
\section*{A.1 Dataset Details}
The dataset details are as follows.

\begin{table}[h]
    \centering
    \begin{tabular}{lcccc}
    \toprule
    \textbf{Sequence} & Total Frames & Train Frames & Validatation Frames & Test Frames \\
    \midrule
    Seattle & 41 & 33 & 4 & 4 \\
    Citron & 37 & 30 & 4 & 3 \\
    Parking & 42 & 34 & 4 & 4 \\ 
    Bike & 104 & 83 & 11 & 10 \\ 
    Jogging & 102 & 82 & 10 & 10 \\ 
    Lab & 103 & 82 & 11 & 10 \\ 
    \bottomrule
    \end{tabular}
    \caption{Number of frames in each dataset used for training, validation and test.}
    \label{tab:dataset_detals}
\end{table}

\section*{A.2 SMPL Refinement}
Given an image, we regress the 2D joints $j_{2d}$ and segmentation mask $m$ of the human using HigherHRNet~\cite{cheng2020bottom} and DensePose~\cite{guler2018densepose}. We further estimate the SMPL mesh $M = (V, F)$, a collection of vertices and faces using ROMP~\cite{ROMP}. The mesh $M$ is parametrized by SMPL parameters $\theta$ such that $ M = \text{SMPL}(\theta)$ and includes the 3D joints ${j}_{3d}$. The regressed SMPL parameters $\theta$ are noisy. Therefore, we use soft-rasterizer~\cite{liu2019soft}, $\Pi$ to refine these estimates. Given a mesh, $M$ and camera $\theta_c$, the rasterizer renders a silhouette $\hat{m} = \Pi(\theta_c, M)$. We also project the 3D joints in the image plane using camera matrix $\hat{j}_{2d} = \mathbf{p}(j_{3d})$ where $\mathbf{p}$ is a projection operator. We obtain the refined SMPL parameters and camera estimates by minimizing

\begin{equation}
    \theta^* = \min_{\theta} \parallel m - \hat{m} \parallel + \parallel j_{2d} - \hat{j}_{2d} \parallel .
\end{equation}

Notice that we use the estimates from DensePose~\cite{guler2018densepose} as the target silhouettes in the preprocessing, while using the estimates from Mask-RCNN~\cite{he2017mask} as the target masks during the training of human NeRF model. It's because in the preprocessing phase, the rendered masks are from naked SMPL mesh and DensePose~\cite{guler2018densepose} is trained on such dense SMPL correspondences. However, we wish to learn extra geometry details beyond the SMPL model with the human NeRF model, and Mask-RCNN~\cite{he2017mask} better estimates those 2D details such as the hair and clothes.

\section*{A.3 Network Architecture}
\begin{figure}[H]
\centering
\includegraphics[width=\linewidth]{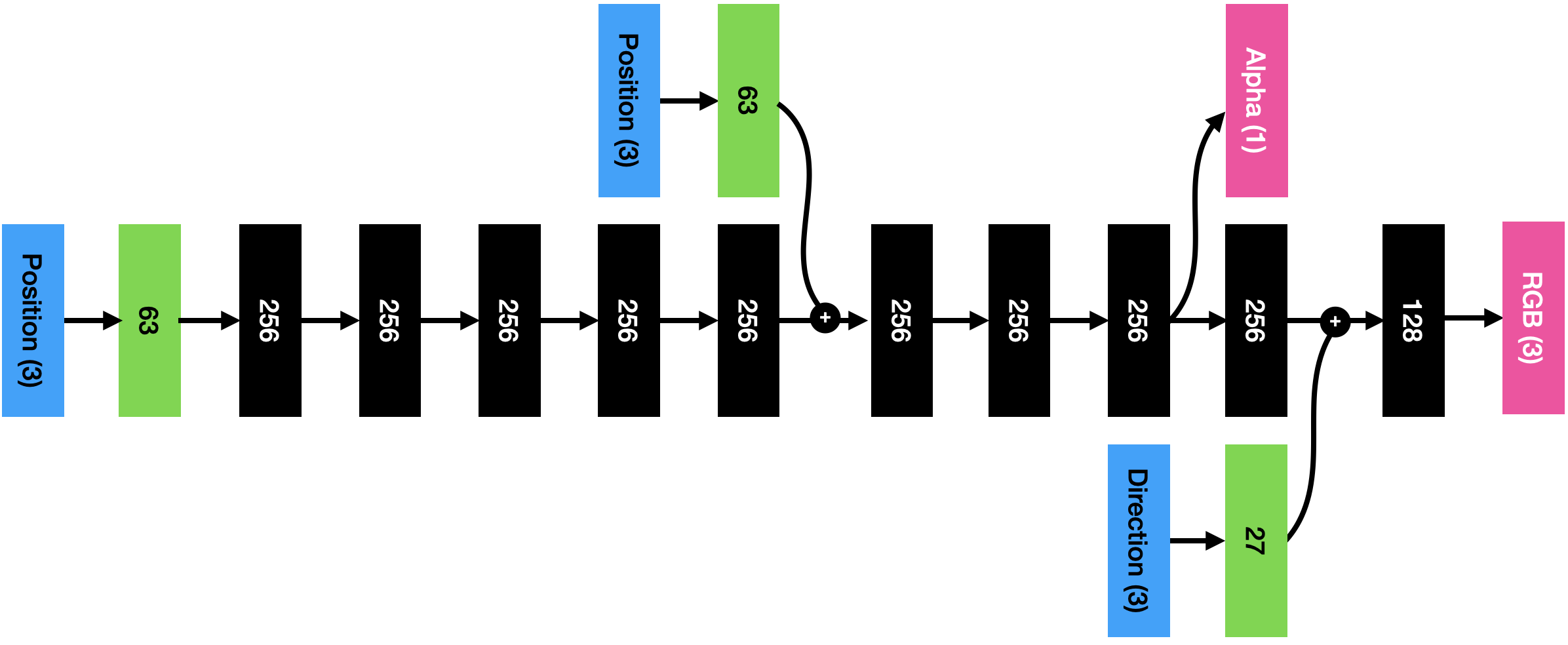}
\caption{
{\bf Visualization of NeRF architecture --}
Blue boxes denote the raw inputs to the network, green boxes denotes the positional encoded values, black boxes denotes the hidden layers, and pink boxes denote the final outputs. The number in each box indicates the feature dimensions. $+$ sign indicates the concatenation operations. ReLU is applied after each hidden layer.
}
\label{fig:arch}
\end{figure}
 
Following~\cite{mildenhall2020nerf}, our scene NeRF models consists of a coarse sub-model and a fine sub-model.
However, with the prior geometry provided by SMPL estimates, we only use one sub-model for the human NeRF model.
Each sub-model and the error-correction network has the same architecture as shown in~\ref{fig:arch}.

\section*{A.4 Comparison with previous works}
\begin{figure}[H]
\centering
\includegraphics[width=0.9\linewidth]{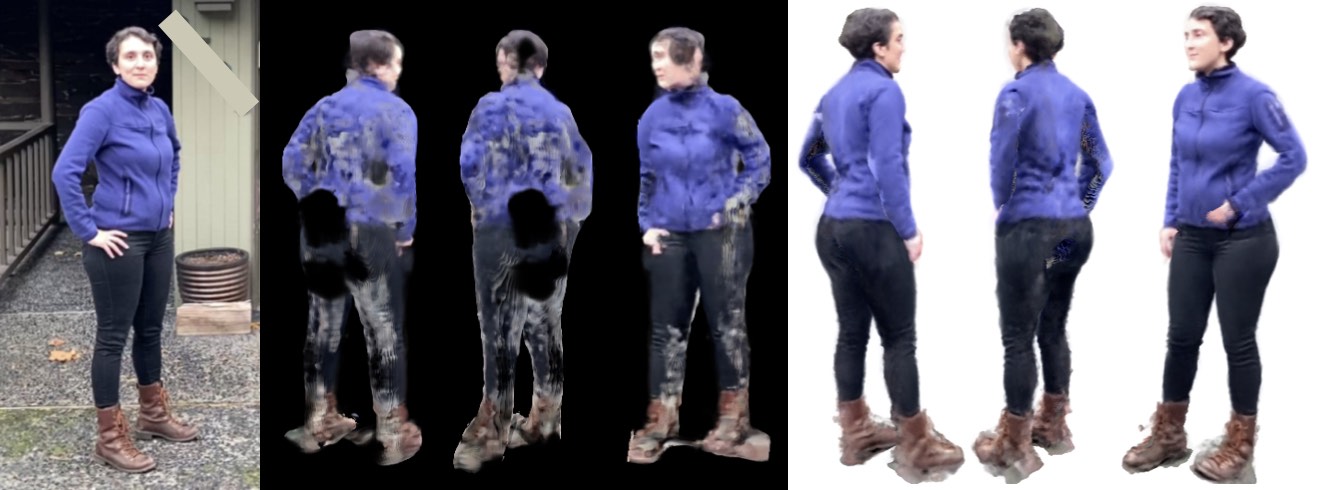}
\caption{
{\bf Novel view rendering of NeuralBody vs Ours on Seattle sequence. --}
The pose being rendered is from the left training image. Three images in the middle are the novel view renderings from NeuralBody, and three images on the right are from ours.
}
\label{fig:nb_v_ours}
\end{figure}
 
We apply NeuralBody~\cite{peng2021neuralbody} to our dataset in a monocular setting. The results are shown in~\ref{fig:nb_v_ours}.
NeuralBody~\cite{peng2021neuralbody} overfits to the training observations, and produce poor rendering on the back of the subject, while ours generalize better and can faithfully render the back.

\begin{figure}[H]
\centering
\includegraphics[width=0.9\linewidth]{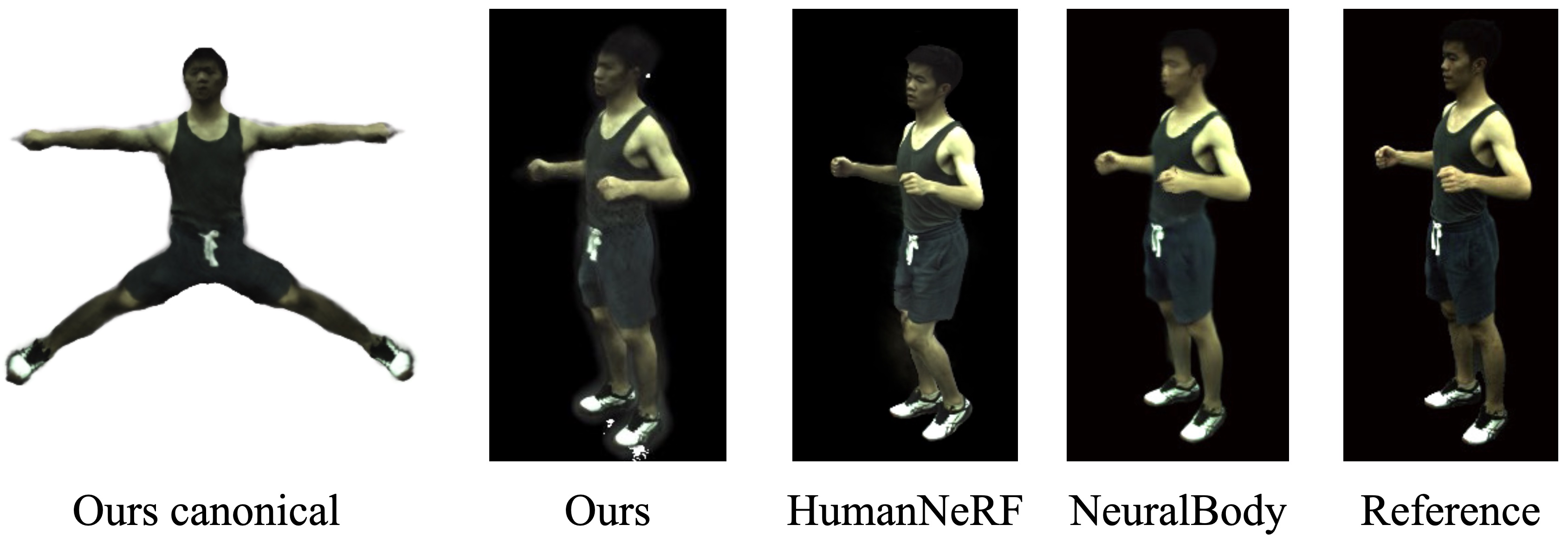}
\caption{
{\bf Novel View Reconstructions on public ZJU Mocap dataset~\cite{peng2021neuralbody} --}
Ours and HumanNeRF~\cite{weng2022humannerf} use only one camera view, NeuralBody~\cite{peng2021neuralbody} uses multiple camera views.
}
\vspace{-1em}
\label{fig:ours_zju_mocap}
\end{figure}

We also compare our method with HumanNeRF~\cite{weng2022humannerf} and NeuralBody~\cite{peng2021neuralbody} on a ZJU Mocap dataset, as shown qualitative comparisons in Figure~\ref{fig:ours_zju_mocap}. Our method renders high quality novel view renderings with the ability to extrapolate in pose space.

\section*{A.5 Error Correction Network and Scene Model Conditioning}
\begin{figure}[h]
\centering
\includegraphics[width=0.9\linewidth]{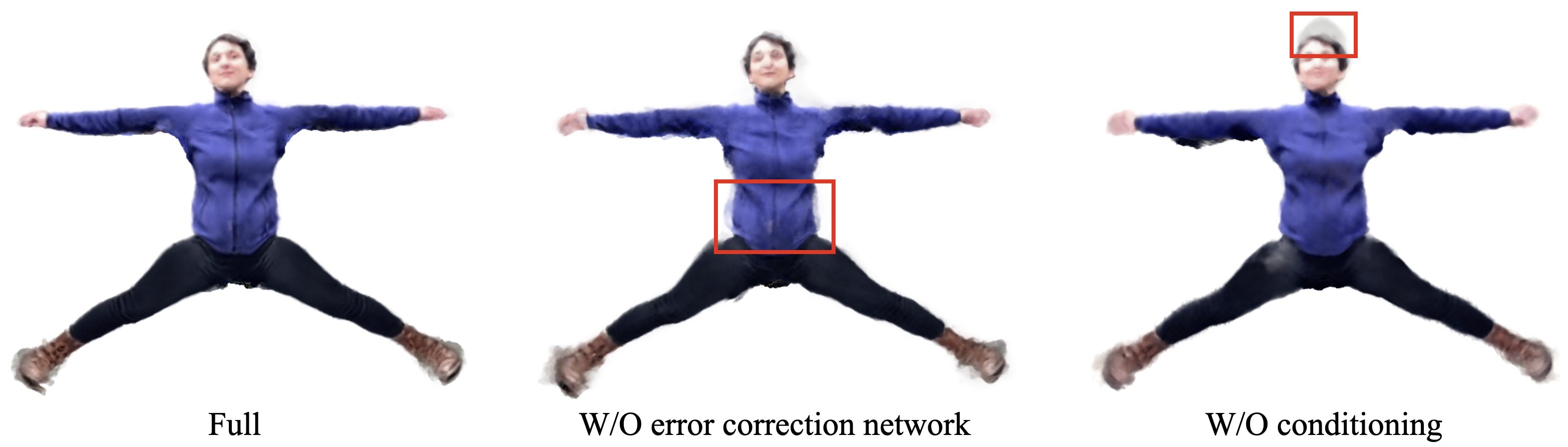}
\vspace{-1em}
\caption{
 Canonical renderings of our full model, model without error correction network, and model without conditioning on scene NeRF.
}
\vspace{-1.5em}
\label{fig:models_comp}
\end{figure}

Without the error correction network, the the canonical NeRF lacks details on the cloth and face.
Training only the human NeRF in isolation leads to worse performance as the human NeRF model may encode the background pixels into its radiance fields due to segmentation errors.
In either case, the canonical NeRF creates fogs around the human to hallucinate the clothing dynamics or the background colors, as shown in ~\ref{fig:models_comp}.

\end{CJK*}
\end{document}